\documentclass[11pt]{article}

\usepackage{amsmath, amsfonts, amssymb, amsthm}
\usepackage{graphicx}
\graphicspath{{./}}
\usepackage{hyperref}
\usepackage{geometry}
\usepackage{authblk}
\usepackage{algorithm2e}
\usepackage{listings}
\usepackage{xcolor}
\usepackage[numbers]{natbib}
\usepackage{tikz}
\usetikzlibrary{positioning,arrows.meta}
\usepackage{float}
\usepackage{booktabs}
\usepackage{cleveref}

\newtheorem{definition}{Definition}[section]
\newtheorem{theorem}{Theorem}[section]
\newtheorem{proposition}{Proposition}[section]
\newtheorem{corollary}[theorem]{Corollary}


\ifdefined\pdfsuppresswarningdupdest
\fi

\hypersetup{colorlinks=true, linkcolor=blue, citecolor=blue, urlcolor=blue}

\title{\textbf{Variational Phasor Circuits for Phase-Native Brain-Computer Interface Classification}}

\author{
  Dibakar Sigdel\textsuperscript{1}\thanks{devdeep137@gmail.com} \\
  \textsuperscript{1} Mindverse Computing LLC, WA 98087
}

\date{\today}

\begin{document}

\maketitle

\begin{abstract}
We present the Variational Phasor Circuit (VPC), a deterministic classical learning architecture on the continuous $S^1$ unit-circle manifold. Inspired by variational quantum circuits, VPC replaces dense weight matrices with trainable phase shifts, local unitary mixing, and structured interference in the ambient complex space, giving a unified method for binary and multi-class classification of spatially distributed signals. We evaluate VPC on real motor-imagery electroencephalography (EEG) from the PhysioNet Motor Movement/Imagery database (10 subjects, Common Spatial Pattern features, subject-wise cross-validation), where it attains a mean decoding accuracy of $0.60$---the highest among standard brain--computer-interface baselines (linear discriminant analysis, logistic regression, RBF-SVM, and a multilayer perceptron)---using an order of magnitude fewer parameters and the lowest cross-subject variance. We also characterize capacity honestly: with phase-only shifts and unitary mixing, VPC realizes a linear decision function in a fixed cosine/sine feature lifting, well matched to the largely separable band-power structure of EEG but unable to represent parity-type functions, a ceiling that depth does not raise. These results position unit-circle phase interference as a parameter-efficient alternative to dense neural computation for signal classification, and motivate VPC both as a standalone classifier and a front-end for hybrid phasor--quantum systems.
\end{abstract}

\section{Introduction}

Electroencephalography (EEG) and brain--computer interface (BCI) research have long served as demanding benchmarks for machine learning because mental-state decoding requires the discrimination of weak, noisy, spatially correlated, and strongly non-stationary signals. The classical BCI literature has therefore explored a wide range of classifiers, from linear discriminants and support vector machines to ensemble methods and multilayer neural networks, each balancing robustness, calibration, and computational cost in different ways \cite{lotte2018review}. At the same time, the underlying neurophysics of EEG is intrinsically oscillatory: signals are generated by coupled cortical sources whose observable structure depends not only on amplitude but also on synchronization, phase relationships, and distributed field interactions across electrodes \cite{nunez2006electric}. This makes BCI classification a natural setting in which to ask whether phase-native learning architectures can offer a better inductive bias than standard Euclidean feature processing.

Historically, foundational BCI systems established the communication-and-control paradigm and emphasized practical trade-offs among signal quality, model complexity, and real-time reliability \cite{wolpaw2002brain}. More recent reviews document a methodological progression from handcrafted spectral/spatial features and shallow classifiers to increasingly deep representation-learning pipelines \cite{lotte2018review}. In parallel, the theory of coupled oscillators has provided a rigorous language for synchronization, phase-locking, and collective dynamics in complex systems \cite{kuramoto1975,strogatz2000kuramoto}, which aligns naturally with phase-centric perspectives on neural data. These two threads---practical BCI machine learning and phase-dynamics theory---motivate architectures that treat phase geometry as a first-class computational object rather than as a derived feature.

Most contemporary machine-learning pipelines still operate through dense real-valued parameter matrices. Whether the model is a shallow multilayer perceptron or a larger attention-based architecture, the dominant computational primitive remains the learned affine map in $\mathbb{R}^N$. Such models are flexible, but their parameter count and memory footprint scale rapidly with feature dimension, and they do not natively enforce the angular or interference structure that often characterizes oscillatory biosignals. In practice, this means that representations of phase-coherent EEG states are frequently mediated by over-parameterized real-valued layers whose geometry is only indirectly related to the signal domain itself.

Quantum machine learning offers an alternative viewpoint. Parameterized and variational quantum circuits replace dense Euclidean weight matrices with sequences of unitary operations acting on encoded states in Hilbert space \cite{benedetti2019parameterized, cerezo2021variational}. This line of work has made parameter efficiency and data encoding central design questions, since the expressive power of the model depends strongly on how classical information is loaded into phase and amplitude coordinates \cite{schuld2021effect}. Yet practical deployment remains constrained by noisy intermediate-scale quantum hardware, limited qubit counts, and difficult optimization phenomena such as barren plateaus \cite{mcclean2018barren}. These limitations motivate a classical analogue that retains the useful unitary, phase-centric structure of variational circuits without inheriting the hardware burden of current quantum devices.

Within the VQC literature itself, multiple formulations have reinforced this perspective: circuit-centric quantum classifiers formalize model families around explicit unitary templates \cite{schuld2020circuit}, hybrid variational theory clarifies optimization principles for parameterized quantum-classical workflows \cite{mcclean2016theory}, and quantum circuit learning demonstrates how trainable unitary stacks can perform supervised function approximation \cite{mitarai2018quantum}. Taken together, these results suggest that structured unitary computation is a powerful inductive bias even before fault-tolerant quantum hardware is widely available.

That analogue is provided by the broader PhasorFlow framework \cite{sigdel2026phasorflow}, which formalizes computation on the unit circle $S^1$ and on the torus $\mathbb{T}^N$ using deterministic phasor states, unitary interference operators, and explicit pull-back operations. Within that framework, the \textbf{Variational Phasor Circuit} (VPC) can be understood as a phase-native classical learning model: data are encoded as unit-magnitude complex states $z_k = e^{i\phi_k}$, trainable parameters enter only as phase shifts, and global structure is induced through local mixing and spectral interference rather than through dense learned Euclidean matrices. The result is a deterministic architecture whose trainable complexity scales linearly with the number of computational threads, while still supporting non-trivial interference dynamics in the ambient space $\mathbb{C}^N$.

This phase-native formulation is especially relevant for EEG and BCI. Because electrode snapshots can be mapped directly to bounded phase coordinates, the encoder preserves the distributed spatial organization of the measurement rather than flattening it immediately into an unconstrained real-valued feature vector. The same bounded representation also makes VPC a plausible \emph{buffer} or interface layer for future hybrid quantum workflows: a classical phasor front-end can absorb raw BCI variability, stabilize it on $\mathbb{T}^N$, and pass structured phase coordinates to downstream quantum or variational quantum models when such hardware becomes practical. In that sense, VPC is not only a standalone classifier but also a candidate preprocessing and encoding system for future phasor-to-qubit pipelines, where careful classical data encoding is already known to be a decisive factor in quantum-model performance \cite{schuld2021effect}.

In this paper, we develop the mathematical construction of VPC and test it on a synthetic 32-channel BrainFlow-based BCI benchmark \cite{brainflow}. The task is designed around binary and four-class mental-state decoding, where the classes differ by distributed spatial phase structure rather than by a simple scalar threshold. Our objective is not merely to show that VPC can classify EEG-like data, but to establish why it is significant: it offers a phase-aligned, low-parameter, fully deterministic alternative to dense neural classifiers, and it provides a concrete computational bridge between classical oscillatory signal processing and future hybrid quantum systems.

The main contributions of this manuscript are threefold:
\begin{enumerate}
	\item We formalize the Variational Phasor Circuit as a deterministic, unit-circle learning architecture derived from the PhasorFlow framework, clarifying its relation to variational quantum circuits while remaining fully classical.
	\item We position VPC in the context of BCI and EEG literature as a phase-native model for mental-state classification, motivated by the oscillatory and spatially distributed structure of neural measurements.
	\item We demonstrate empirically that shallow and depth-scaled VPC models can achieve strong BCI classification performance with drastically fewer trainable parameters than standard dense baselines, while also serving as a plausible front-end encoding layer for future phasor--quantum pipelines.
\end{enumerate}

\section{Theory}

This section develops VPC in four components: (i) phasor state geometry on $\mathbb{T}^N$, (ii) unitary gate primitives in the ambient space $\mathbb{C}^N$, (iii) single-block VPC operators, and (iv) deep composition with inter-block pull-back.

\subsection{Phasor State Vectors}

Unlike a classical computer which manipulates discrete binary strings $b \in \{0, 1\}^N$, or a quantum computer operating on complex probability amplitudes $|\psi\rangle$, a Phasor computer operates entirely on continuous angular states in the interval $[-\pi, \pi]$. Each computational thread is a unit-magnitude complex number identified with $U(1)$:
\begin{equation}
z = e^{i\phi}, \qquad \phi\in(-\pi,\pi].
\end{equation}
For $N$ threads, the encoded state is
\begin{equation}
\boldsymbol{z} = \bigl(e^{i\phi_1},\dots,e^{i\phi_N}\bigr)^\top \in \mathbb{T}^N \subset \mathbb{C}^N,
\end{equation}
where $\phi_k=\arg(z_k)\in(-\pi,\pi]$ is the phase coordinate of thread $k$ (principal branch).

\begin{definition}[VPC State Manifold]
For $N$ computational threads, the admissible state manifold is
\begin{equation}
\mathcal{M}_{\mathrm{VPC}} = \mathbb{T}^N = \{\mathbf{z}\in\mathbb{C}^N:\ |z_k|=1,\;k=1,\dots,N\}.
\end{equation}
We reserve $\phi$ for phasor-state angles and use $\theta$ for trainable gate parameters.
\end{definition}

\begin{definition}[Ambient Complex Extension]
Although admissible phasor states lie on $\mathbb{T}^N$, linear interference operators act on the ambient vector space $\mathbb{C}^N$. Hence a unitary map $U\in U(N)$ preserves global $\ell^2$ energy but need not preserve coordinatewise unit modulus.
\end{definition}

\subsection{Unitary Gate Primitives}

The VPC block uses two unitary operator classes:
\begin{equation}
S(\boldsymbol{\theta})=\mathrm{diag}\!\left(e^{i\theta_1},\dots,e^{i\theta_N}\right)\in U(1)^N\subset U(N),
\qquad
M_{j,k}=\frac{1}{\sqrt{2}}\begin{pmatrix}1 & i\\ i & 1\end{pmatrix}\in U(2).
\end{equation}
$S(\boldsymbol{\theta})$ is a coordinatewise phase rotation that preserves $\mathbb{T}^N$ exactly; $M_{j,k}$ is a local beam-splitter coupling adjacent threads $j$ and $k$.

Applying $M_{j,k}$ to $(z_j,z_k)=(e^{i\phi_j},e^{i\phi_k})$ yields $|z_j'|^2=1+\sin(\phi_j-\phi_k)$, so individual thread magnitudes can depart from unity while global $\ell^2$ energy is conserved.

\begin{proposition}[Unitary Mixing Extends VPC States into $\mathbb{C}^N$]
Let $\mathbf{z}\in\mathbb{T}^N$ and let $U\in U(N)$ be a non-diagonal mixing operator. Then
\begin{equation}
\|U\mathbf{z}\|_2 = \|\mathbf{z}\|_2 = \sqrt{N},
\end{equation}
but in general $U\mathbf{z}\notin\mathbb{T}^N$ because the coordinatewise modulus constraints $|(U\mathbf{z})_k|=1$ need not hold.
\end{proposition}

\subsection{Single-Block VPC Operator}
Given encoded input state $\mathbf{z}_{\mathrm{in}}\in\mathbb{T}^N$, a single VPC block is
\begin{equation}
\mathcal{V}(\boldsymbol{\theta}) = U_{\mathrm{local}}\left(\prod_{k=1}^{N} S_k(\theta_k)\right),
\qquad
U_{\mathrm{local}}=\prod_{k=0,2,4,\ldots}M_{k,k+1},
\end{equation}
with forward state $\mathbf{z}_f=\mathcal{V}(\boldsymbol{\theta})\mathbf{z}_{\mathrm{in}}$.

\subsection{Circuit Schematics for VPC Blocks}

\begin{figure}[htbp]
\centering
\sffamily
\definecolor{cCh3}{HTML}{9C27B0}
\definecolor{cCh2}{HTML}{4CAF50}
\definecolor{cCh1}{HTML}{03A9F4}
\definecolor{cCh0}{HTML}{F44336}
\begin{tikzpicture}[
	scale=0.78, transform shape,
	node distance=1.5cm,
	rail/.style={line width=5pt, line cap=round, opacity=0.8},
	shift_gate/.style={rectangle, draw, fill=white, line width=1.0pt, minimum width=0.6cm, minimum height=0.6cm, font=\small\bfseries, rounded corners=2pt},
	mix_gate/.style={rectangle, draw, fill=gray!10, line width=1.0pt, minimum width=0.4cm, minimum height=1.8cm, font=\small\bfseries, rounded corners=2pt},
	connection/.style={line width=3pt, color=gray!50}
]

	\def\threads{
		3/Thread $z_3$/cCh3,
		2/Thread $z_2$/cCh2,
		1/Thread $z_1$/cCh1,
		0/Thread $z_0$/cCh0%
	}
    
	\foreach \y/\ilabel/\icolor in \threads {
		 \pgfmathsetmacro{\ypos}{\y * 1.8}
		 \draw[rail, color=\icolor] (0, \ypos) -- (13.5, \ypos);
		 \node[anchor=east, color=\icolor, font=\small\bfseries, align=right] at (-0.2, \ypos) {\ilabel};
	}
    
	\node[anchor=east, font=\large\bfseries, align=center] at (-2.0, 2.7) {Input State\\[0.5ex] $\boldsymbol{z}_{\mathrm{in}}$};
	\draw[->, line width=2pt, color=gray] (-1.8, 2.7) -- (-0.5, 2.7);

	\foreach \y in {0,1,2,3} {
		 \pgfmathsetmacro{\ypos}{\y * 1.8}
		 \node[shift_gate, draw=black] at (2.0, \ypos) {$S(\theta^{\text{in}}_{\y})$};
	}
    
	\draw[connection] (4.0, 5.4) -- (4.0, 3.6);
	\node[mix_gate, minimum height=2.0cm] at (4.0, 4.5) {$M_{2,3}$};
	\draw[connection] (4.0, 1.8) -- (4.0, 0.0);
	\node[mix_gate, minimum height=2.0cm] at (4.0, 0.9) {$M_{0,1}$};

	\foreach \y in {0,1,2,3} {
		 \pgfmathsetmacro{\ypos}{\y * 1.8}
		 \node[shift_gate, draw=black] at (6.0, \ypos) {$S(\theta_{\y})$};
	}

	\draw[connection] (8.0, 3.6) -- (8.0, 1.8);
	\node[mix_gate, minimum height=2.0cm] at (8.0, 2.7) {$M_{1,2}$};

	\foreach \y in {0,1,2,3} {
		 \pgfmathsetmacro{\ypos}{\y * 1.8}
		 \node[shift_gate, draw=black] at (10.0, \ypos) {$S(\theta_{\y}^{\text{out}})$};
	}

	\draw[->, line width=2pt, color=gray] (11.7, 2.7) -- (12.7, 2.7);
	\node[anchor=west, font=\large\bfseries, align=center] at (12.8, 2.7) {Output State\\[0.5ex] $\boldsymbol{z}_f$};
    
\end{tikzpicture}
\caption{Single-stack Variational Phasor Circuit (VPC) schematic demonstrating consecutive shift--mix topologies. All interior shifts ($\theta$) represent pure trainable parameters; the phase-domain data encoding occurs entirely outside the bounds of the VPC block.}
\label{fig:vpc_arch}
\end{figure}

\begin{figure}[H]
\centering
\begin{tikzpicture}[
	node distance=2.2cm,
	block/.style={rectangle, draw, fill=blue!10, text width=2.5cm, text centered, rounded corners, minimum height=1.0cm, font=\small\bfseries},
	layerblock/.style={rectangle, draw, fill=green!10, text width=2.6cm, text centered, rounded corners, minimum height=0.9cm, font=\small},
	arrow/.style={thick,->,>=stealth}
]

\node (input) {Input Features $\boldsymbol{r} \in \mathbb{R}^d$};
\node (encode) [block, below of=input, node distance=2.0cm] {Phase Encoding\\$\boldsymbol{r}\rightarrow\boldsymbol{\phi}$};

\node (vpc_layer) [rectangle, draw, dashed, fill=gray!5, minimum width=8.5cm, minimum height=6.5cm, below of=encode, node distance=4.5cm] {};
\node [anchor=north west, font=\small\bfseries] at (vpc_layer.north west) {VPC Layer $\times L$};

\node (shift) [layerblock, below of=encode, node distance=2.3cm] {Shift Gate\\ (Parameters $\boldsymbol{\theta}$)};
\node (mix) [layerblock, below of=shift, node distance=2.0cm] {Mix Gate\\ (Coupling)};
\node (thresh) [layerblock, below of=mix, node distance=2.0cm] {Normalize Gate\\ (Manifold Pull-Back)};

\node (output) [below of=thresh, node distance=2.2cm] {Output State $\boldsymbol{z}_f$};

\draw [arrow] (input) -- (encode);
\draw [arrow] (encode) -- (shift);
\draw [arrow] (shift) -- (mix);
\draw [arrow] (mix) -- (thresh);
\draw [arrow] (thresh) -- (output);

\draw [arrow, dashed] (thresh.east) -- ++(1.5,0) |- node[anchor=west, pos=0.25] {Repeat $L$ times} (shift.east);

\end{tikzpicture}
\caption{Deep-stack VPC schematic where each block is separated by pull-back normalization for stable depth scaling.}
\label{fig:vpc_multistack_arch}
\end{figure}

\begin{theorem}[Pull-Back Stabilization for Deep VPC Cascades]
Consider a depth-$S$ VPC cascade of $S$ stacked blocks, where each linear block is composed of Shift/Mix operators and each inter-block transition applies threadwise pull-back
\begin{equation}
\mathcal{P}(z_k)=\frac{z_k}{|z_k|},\quad z_k\neq 0.
\end{equation}
Then every nonzero thread entering each subsequent block satisfies unit modulus, i.e., $|\mathcal{P}(z_k)|=1$. Consequently, deep composition preserves phase coordinates on $\mathbb{T}^N$ at block boundaries and prevents unbounded amplitude propagation across layers.
\end{theorem}

\subsection{Analytical Forward Pass and Readout}
For a single-stack VPC with one variational layer, the forward map can be written as
\begin{equation}
\boldsymbol{z}_f = U_{\mathrm{local}}
\left(\prod_{k=1}^{N} S_k(\theta_k)\right)
\boldsymbol{z}_{\mathrm{in}}.
\label{eq:vpc_theory_forward_l1}
\end{equation}
where $\boldsymbol{z}_{\mathrm{in}}=(e^{i\phi_1},\ldots,e^{i\phi_N})^\top$ and $U_{\mathrm{local}}=\prod_{k=0,2,\ldots}M_{k,k+1}$.

For binary decoding, a single output thread angle can be mapped to probability by
\begin{equation}
P(y=1\mid\mathbf{z}_{\mathrm{in}},\boldsymbol{\theta}) = \frac{\sin(\phi_0)+1}{2},\qquad \phi_0=\arg(z_{f,0}).
\end{equation}
For multi-class categorical classification across $K$ classes, the VPC collapses the complex state space into mutually exclusive probability assignments by observing the first $K$ computational threads. Writing the principal phase of thread $k$ as
\begin{equation}
\phi_k = \arg(z_{f,k}) \in(-\pi,\pi],
\end{equation}
we define the class logit through a bounded sinusoidal envelope with fixed scale $s>0$,
\begin{equation}
L_k = s\,\sin(\phi_k), \qquad k=1,\dots,K,
\end{equation}
which are then passed through a standard softmax bottleneck,
\begin{equation}
P(y=k\mid\mathbf{z}_{\mathrm{in}},\boldsymbol{\theta}) = \frac{e^{\,s\sin\phi_k}}{\sum_{j=1}^{K} e^{\,s\sin\phi_j}}.
\end{equation}
The $\sin(\cdot)$ envelope makes each logit a smooth, bounded function of the thread phase across the periodic boundary $\phi=\pm\pi$, keeping the cross-entropy landscape differentiable under Autograd; the fixed scale $s$ (implementation default $s=5$) sets the sharpness of the posterior. The trainable phase parameters $\boldsymbol{\theta}$ are optimized against categorical cross-entropy with the Adam optimizer over the ambient complex geometry.

\subsection{From Single-Stack to Multi-Stack VPC}
\label{subsec:single_to_multi_stack_vpc}
When scaling the Variational Phasor Circuit to increasingly complex topological classification tasks, there are two distinct architectural paradigms for increasing representational depth: the \emph{Deep Circuit} and the \emph{Deep Stack}.

In the Deep Circuit scenario, depth is increased by continuously appending additional parameterized Shift gates and unitary Mix gates without intermediate geometric correction. Because there is no non-linear renormalization, the entire circuit remains a purely continuous unitary cascade in the ambient space $\mathbb{C}^N$. The state vector therefore propagates through uninterrupted linear wave interference. As depth grows, constructive and destructive interference naturally drive individual thread magnitudes $|z_k|$ away from the torus manifold $\mathbb{T}^N$ and deeper into the unconstrained complex amplitude space. This increases representational richness, but the resulting amplitude excursions can become statistically unstable and difficult to optimize at large depth.

With inter-block pull-back, an $S=2$ deep-stack form (two stacks) is
\begin{equation}
\boldsymbol{z}_f = \mathcal{P}\!\left(U_{\mathrm{local}}^{(2)}\prod_k S_k(\theta_k^{(2)})\;\mathcal{P}\!\left(U_{\mathrm{local}}^{(1)}\prod_k S_k(\theta_k^{(1)})\,\boldsymbol{z}_{\mathrm{in}}\right)\right).
\label{eq:vpc_theory_forward_l2}
\end{equation}
Conversely, in the Deep Stack architecture, multiple isolated VPC blocks are composed sequentially with a non-linear Normalize (pull-back) gate inserted between them. This operation rigidly normalizes the scalar magnitude of each nonzero complex thread back to $1$, explicitly pulling dispersed phasors back onto the torus manifold before another round of unitary mixing begins. By severing runaway amplitude excursions while preserving the encoded phase information, the Deep Stack permits substantially deeper compositions than a raw uninterrupted circuit.

This expresses the key structural distinction: deep stacks introduce geometric stabilization boundaries between otherwise unitary interference blocks. Each layer can therefore process fresh phase interactions without being overwhelmed by amplitude resonance inherited from previous layers.

\begin{proposition}[Linear Parameter Footprint of Deep VPC]
Consider a VPC with $N$ channels, $L$ variational shift layers per stack, and $S$ stacks, where each layer contributes one phase parameter per channel. The total trainable parameter count is
\begin{equation}
P_{\mathrm{VPC}} = N\,L\,S.
\end{equation}
Therefore scaling channel count, layer count, or stack count increases parameters linearly, while the Mix and DFT topology remains parameter-free. The single-stack case ($S=1$) reduces to $P_{\mathrm{VPC}}=NL$; the configurations reported in this paper use $L=2$, so a $1,2,3,4$-stack model has $2N,4N,6N,8N$ parameters respectively.
\end{proposition}

\begin{corollary}[Efficiency Regime of Deep-Stack VPC]
Combining pull-back stabilization with linear parameter scaling, deep-stack VPC architectures define a regime where high classification performance can be achieved with substantially fewer trainable parameters than dense Euclidean baselines, while preserving explicit phase-manifold interpretability.
\end{corollary}

\section{Method}

This section describes the implementation protocol used for VPC experiments. Operator-level derivations and formal statements are provided in Theory; here we focus on the practical workflow needed for reproducible training and evaluation. We use two data regimes. The \emph{primary} regime is real motor-imagery EEG from the PhysioNet EEG Motor Movement/Imagery database (\Cref{subsec:real_eeg}), which provides the substantive evidence for the model's practical value. A \emph{controlled synthetic} regime is used only for a training-behavior check and for the capacity probe, where knowing the ground-truth separability of the task is essential. In all regimes the pipeline is the same four stages: (1) data are prepared, labeled, and split; (2) each feature vector is normalized and mapped to bounded phase coordinates and lifted onto the phasor manifold; (3) single- or multi-stack VPC models are fit under fixed optimization budgets using PyTorch Autograd; and (4) performance is measured on held-out data and compared against classical baselines under identical splits. All splitting for the real-EEG regime is subject-wise stratified $k$-fold cross-validation to prevent trial leakage and to report generalization across individuals.

\subsection{End-to-End Experimental Workflow}

\begin{figure}[htbp]
\centering
\begin{tikzpicture}[
    scale=0.85, transform shape,
    node distance=1.8cm,
    block/.style={rectangle, draw, fill=blue!10, text width=3.5cm, text centered, rounded corners, minimum height=1.0cm, font=\small\bfseries},
    layerblock/.style={rectangle, draw, fill=orange!10, text width=4.5cm, text centered, rounded corners, minimum height=0.9cm, font=\small},
    arrow/.style={thick,->,>=stealth}
]

\node (input) {Raw Data $\mathbf{x} \in \mathbb{R}^{N}$};

\node (norm) [block, below of=input, node distance=1.5cm] {Normalize \\ $[-\pi/2, \pi/2]$};
\node (encode) [block, below of=norm, node distance=1.5cm] {Phase Encode \\ $\mathbf{z}=e^{i\boldsymbol{\phi}}$};

\node (vpc_layer) [rectangle, draw, dashed, fill=gray!5, minimum width=7.0cm, minimum height=5.5cm, below of=encode, node distance=3.7cm] {};
\node [anchor=north west, font=\small\bfseries] at (vpc_layer.north west) {VPC Blocks (Repeated $S$ times)};

\node (shift) [layerblock, below of=encode, node distance=2.2cm] {Phase Shift $\leftarrow S(\boldsymbol{\theta})$};
\node (mix) [layerblock, below of=shift, fill=red!10, node distance=1.5cm] {Local Mixing $\leftarrow M$};
\node (pullback) [layerblock, below of=mix, fill=green!10, node distance=1.5cm] {Pull-back Normalize $\leftarrow \mathcal{P}$};

\node (readout) [below of=pullback, node distance=2.0cm, font=\small\bfseries] {Readout (Class Logits)};

\node (loss) [rectangle, draw, dashed, thick, fill=red!10, right of=vpc_layer, node distance=8.0cm, text width=3.0cm, text centered, rounded corners, minimum height=1.2cm, font=\small\bfseries] {Loss + Backprop \\ $\nabla_{\boldsymbol{\theta}}$};

\draw [arrow] (input) -- (norm);
\draw [arrow] (norm) -- (encode);
\draw [arrow] (encode) -- (shift);
\draw [arrow] (shift) -- (mix);
\draw [arrow] (mix) -- (pullback);
\draw [arrow] (pullback) -- (readout);

\draw [arrow, dashed, color=red, thick] (readout.east) -| node[pos=0.25, below, text=black, font=\small] {Compute Loss} (loss.south);
\draw [arrow, dashed, color=red, thick] (loss.west) -- node[above, text=black, font=\small] {$\nabla_{\boldsymbol{\theta}}$ Updates} (vpc_layer.east);

\end{tikzpicture}
\caption{Experimental workflow used in VPC training and evaluation, from raw data preprocessing to phasor inference and gradient-based parameter updates. Phase encoding lifts each channel snapshot onto $\mathbb{T}^N$; each of the $S$ stacked VPC blocks applies a trainable shift layer, local beam-splitter mixing, and pull-back normalization before readout.}
\label{fig:bci_workflow}
\end{figure}

Figure~\ref{fig:bci_workflow} summarizes the end-to-end pipeline. The boundary between preprocessing (Stage~1--2) and the learnable circuit (Stage~3) is architecturally clean: the encoding stage is a fixed, parameter-free lifting operation, while all trainable degrees of freedom reside exclusively in the shift-gate parameters $\boldsymbol{\theta}$ of the VPC blocks. This separation ensures that observed performance differences across experiments can be attributed unambiguously to circuit design and optimization strategy rather than to data-handling choices.

\subsection{Data Generation and Splitting}

Controlled experiments use BrainFlow-based synthetic EEG streams with $N=32$ channels whose spatial patterns are conditioned on the target class label. The primary evaluation setting is a four-class classification problem whose classes represent qualitatively distinct topological regimes: a resting-state pattern with spatially uniform phases, two motor-imagery-like states with lateralized left-hemisphere and right-hemisphere activity profiles, and a globally coherent flow state whose phase distribution spans the full torus uniformly. This design ensures that the classification boundary is genuinely topological rather than a simple amplitude threshold, stressing the phasor manifold representation rather than a scalar feature separator.

For binary-classification ablations, the same generation framework is used with only two active class labels, providing a lower-complexity baseline for assessing parameter scaling. Both settings use fixed random seeds to guarantee reproducible splits. Data are partitioned into training, validation, and held-out test subsets with stratified class balance so that every split contains equal proportions of each mental-state label, preventing any marginal class frequency from confounding the accuracy estimates.

\subsection{Preprocessing and Phase Encoding}

Each acquired sample is a snapshot vector $\mathbf{x} \in \mathbb{R}^N$ representing simultaneously recorded activity across all $N=32$ channels at a single time point. Prior to phase encoding, each snapshot is standardized by subtracting the sample mean and dividing by the sample standard deviation, producing a zero-mean, unit-variance vector $\mathbf{x}_{\mathrm{norm}}$. Phase encoding then maps each normalized channel value to a bounded angular coordinate via the saturating nonlinearity
\begin{equation}
\phi_k = \pi\,\tanh\!\left(x_{\mathrm{norm},k}\right), \qquad k = 1, \ldots, N.
\end{equation}
Because $\tanh$ maps $\mathbb{R}$ continuously and monotonically into $(-1, 1)$, the resulting angles are confined to the open interval $(-\pi, \pi)$, avoiding the degenerate phase wrap-around singularity at $\pm\pi$. The encoded input state is then
\begin{equation}
\boldsymbol{z}_{\mathrm{in}} = \bigl(e^{i\phi_1}, \ldots, e^{i\phi_N}\bigr)^\top \in \mathbb{T}^N,
\end{equation}
which places every sample precisely on the phasor state manifold $\mathcal{M}_{\mathrm{VPC}}$ as defined in the Theory section. This encoding is applied identically across all experiment types---single-stack, deep-stack, and baseline benchmarking---so that observed performance differences reflect circuit architecture and optimization rather than any asymmetry in input representation.

\subsection{Model Construction and Training}

\subsubsection{Single-Stack and Deep-Stack Configurations}

Two VPC architectures are evaluated, each exercising a different regime of the parameter-efficiency trade-off. The single-stack configuration consists of one VPC block---a shift layer $S(\boldsymbol{\theta})$ followed by nearest-neighbour beam-splitter couplings $M_{j,k}$---and serves as the primary model. Its parameter count follows $P_{\mathrm{VPC}}=N\,L$ (Theory): with $L=2$ shift layers, the real-EEG model over $N=6$ CSP features uses $12$ trainable phases, and the synthetic two-state demonstration over $N=32$ channels uses $64$. In both cases the parameter budget is one to two orders of magnitude smaller than the classical baselines against which it is compared.

The deep-stack configuration extends the single-stack by stacking $S$ VPC blocks in series. Following each linear block, a threadwise pull-back normalization
\begin{equation}
\mathcal{P}(z_k) = \frac{z_k}{|z_k|}
\end{equation}
restores every thread to unit modulus before it enters the next block. As established in the Pull-Back Stabilization theorem in the Theory section, this inter-block operation prevents unbounded amplitude drift across depth and keeps phase coordinates on $\mathbb{T}^N$ at every block boundary. The deep-stack therefore enables a principled study of depth scaling on the phasor manifold without the numerical instabilities that arise when amplitudes are allowed to grow freely through successive unitary layers.

Binary classification decodes the circuit output through a single designated thread: the probability of the positive class is
\begin{equation}
P(y=1 \mid \boldsymbol{z}_{\mathrm{in}}, \boldsymbol{\theta}) = \frac{\sin(\phi_{f,0}) + 1}{2}, \qquad \phi_{f,0} = \arg(z_{f,0}).
\label{eq:binary_readout_vpc}
\end{equation}
Four-class decoding reads the first four output threads, computes bounded sinusoidal logits $L_k = s\,\sin(\phi_{f,k})$ (fixed scale $s=5$) for $k = 1, \ldots, 4$, and applies a softmax to produce a normalised probability distribution over classes, exactly as defined in the Analytical Forward Pass section of the Theory.

\subsubsection{Optimization Protocol}

Training is carried out in PyTorch with continuous Autograd differentiation over the trainable shift parameters $\boldsymbol{\theta}$. Each training iteration performs a forward pass on a batch of phase-encoded snapshots, computes the task loss---binary cross-entropy for two-class problems and categorical cross-entropy for the four-class setting---back-propagates gradients through the complex-valued operations, and updates $\boldsymbol{\theta}$ via a gradient step. Training and validation losses are logged at every epoch to monitor convergence and detect over-fitting.

Because the phasor state manifold is periodic, the loss landscape can exhibit local minima that are topologically equivalent under $2\pi$ translation but distinct under Euclidean geometry. The bounded sinusoidal readout envelopes (\Cref{eq:binary_readout_vpc} and the multi-class softmax) keep the objective smooth across these periodic boundaries, so all models in this paper are trained end-to-end with the Adam optimizer operating directly on the ambient complex geometry via PyTorch Autograd. This is the sole optimizer used for the reported results; earlier derivative-free experiments (e.g.\ COBYLA) are not required once the readout is made differentiable and are omitted here for a controlled comparison.

\subsection{Evaluation and Baseline Benchmarking}

\subsubsection{Testing Protocol}

All final performance metrics are computed exclusively on the held-out test partition, which is never accessed during training or hyperparameter selection. The primary metric is classification accuracy, supplemented by confusion matrices that reveal any systematic class-level confusion patterns among the four topological mental states. Training and validation loss curves are reported alongside test accuracy to give a complete picture of optimizer convergence and generalization gap. Parameter counts are also tabulated for every model variant to make the efficiency-accuracy trade-off quantitatively explicit.

For the deep-stack ablation, both the deep-circuit model (additional shift and mix layers without inter-block pull-back) and the deep-stack model (same depth with pull-back normalization at each block boundary) are trained under matched data partitions and matched optimization budgets. The only variable that differs between the two conditions is the presence or absence of the $\mathcal{P}$ normalization step, isolating the contribution of pull-back stabilization to classification performance as circuit depth increases.

\subsubsection{Classical Baselines}

To place VPC performance in the context of the broader machine-learning landscape, we train Decision Tree, Random Forest, Support Vector Machine, and Multilayer Perceptron classifiers on the same phase-encoded feature vectors used for VPC training. All baseline models receive identical train and test splits and are evaluated with the same accuracy and confusion-matrix protocol. This design ensures that any observed performance gap---in accuracy or in trainable parameter count---reflects structural differences between the phasor circuit representation and conventional Euclidean classifiers, rather than any asymmetry in data access or evaluation conditions.

Parameter counts for each baseline are reported alongside VPC counts to quantify the parameter-efficiency advantage. The comparison is particularly sharp for the MLP baseline, which requires dense real-valued weight matrices that grow quadratically with hidden-layer width, against the VPC's $N$-parameter shift-gate structure that grows only linearly with the number of computational threads.

\section{Results}

We evaluate the Variational Phasor Circuit on three progressively more demanding settings: (i) a controlled synthetic task that establishes correct training behavior; (ii) a controlled capacity probe that delineates what the VPC can and cannot represent; and (iii) real motor-imagery EEG, which is the primary evidence for the model's practical value. Throughout, we report explicit baselines and honest parameter counts, and we avoid drawing conclusions from tasks whose separability is not established.

\subsection{Controlled Synthetic Task: Training Behavior}
As an initial check that the phase-native circuit trains and generalizes as a standard model, we deployed a single-stack VPC on a synthetic two-state spatial task ($N=32$). Utilizing PyTorch Autograd to track exact wave derivatives through the complex $\mathbb{C}^N$ state, we optimized against Mean Squared Error with Adam ($\lambda=0.1$) over $60$ epochs; the training loss decayed smoothly from $0.34$ to below $10^{-3}$ and the model reached $100\%$ held-out accuracy. We stress that this task is trivially (linearly) separable by construction and serves only to confirm correct end-to-end training behavior, not to demonstrate representational power; the substantive evaluation is on real EEG (\Cref{subsec:real_eeg}).

We stress that this synthetic task is a \emph{training-behavior check, not evidence of non-linear power}: the state templates are deterministic spatial patterns that are linearly separable, and a logistic-regression baseline on the same features solves it essentially perfectly. The earlier framing of this experiment as ``$100\%$ mental-state decoding'' overstated its significance; we retain it only to demonstrate stable convergence, and defer the substantive claims to the capacity probe (\Cref{subsec:capacity}) and the real-EEG benchmark (\Cref{subsec:real_eeg}).

\begin{figure}[H]
    \centering
    \includegraphics[width=1\textwidth]{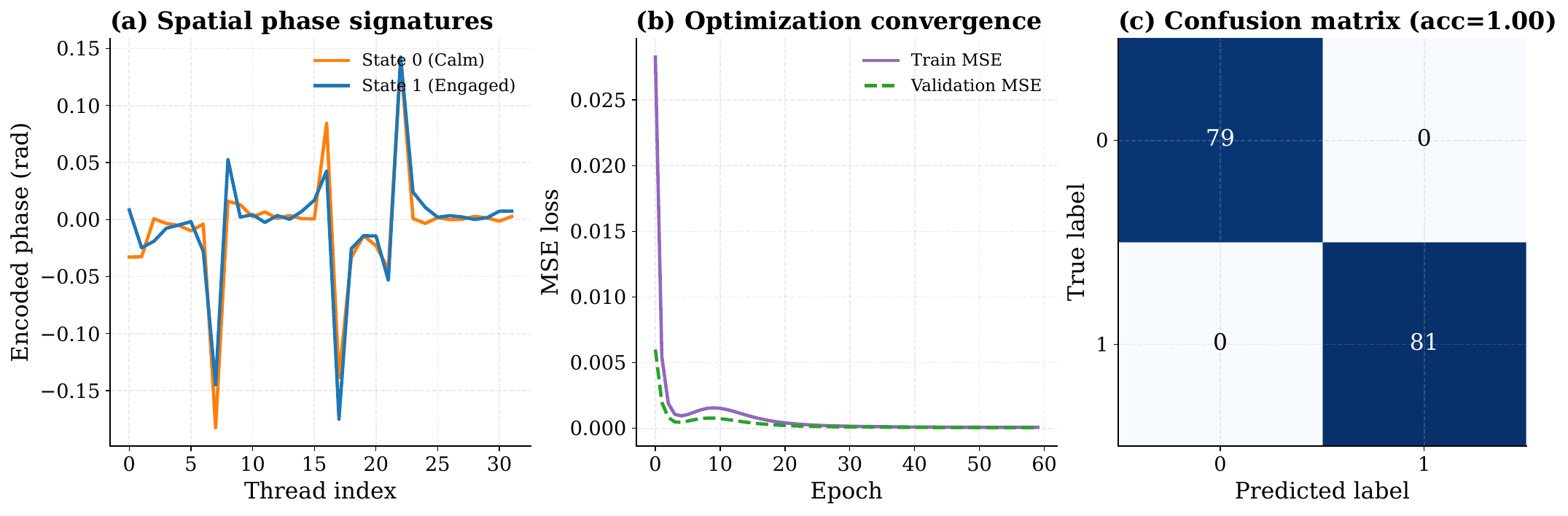}
    \caption{\textbf{Single-stack VPC on the synthetic two-state task ($N=32$).} (A) Phase-space representation of the two synthetic states mapped onto $\mathbb{T}^{32}$. (B) Convergence of the MSE training and validation loss curves under PyTorch Autograd against the exact analytic simulator. (C) Validation confusion matrix. This task establishes correct training behavior only; its separability is trivial (see text).}
    \label{fig:binary_multipanel}
\end{figure}

\subsection{Capacity Probe: The Phase-Linear Ceiling}
\label{subsec:capacity}

To determine what the VPC can represent, we use a controllable phase-parity task: given $N=8$ phase inputs, the label is $y=\big(\sum_{j=1}^{k}\mathbb{1}[\cos x_j>0]\big)\bmod 2$. Parity is the canonical function that is \emph{not} linearly separable, and its order $k$ tunes difficulty. \Cref{tab:capacity} reports test accuracy at $k=2$ (phase-XOR) across depths, with baselines (3-seed mean).

\begin{table}[h]
\centering
\begin{tabular}{|l|c|c|}
\hline
\textbf{Model} & \textbf{Test Accuracy} & \textbf{Trainable Params} \\ \hline
Logistic regression (cos features) & $0.49$ & 17 \\ \hline
VPC, 1 stack & $0.48$ & 16 \\ \hline
VPC, 2 stacks (pull-back) & $0.51$ & 32 \\ \hline
VPC, 3 stacks (pull-back) & $0.49$ & 48 \\ \hline
RBF-SVM & $0.91$ & Non-parametric \\ \hline
MLP (32 hidden) & $0.98$ & 321 \\ \hline
\end{tabular}
\caption{VPC on phase-XOR ($k=2$, $N=8$). The VPC remains at chance at every depth, while a kernel SVM and an MLP solve the task.}
\label{tab:capacity}
\end{table}

The VPC sits at chance on phase-XOR regardless of depth. This is a structural property, not a training failure: with diagonal (phase-only) shift gates, fixed unitary Mix/DFT coupling, and a phase readout, the VPC computes a \emph{linear decision function in a fixed $\cos/\sin$ lifting} of its inputs. It can represent any phase-separable boundary but not multiplicative interactions such as parity, and---crucially---stacking with pull-back re-projection does not raise this ceiling, because projecting back onto $\mathbb{T}^N$ between stacks discards the amplitude interference needed to build non-linear features. This behavior is directly analogous to random-Fourier-feature and reservoir-computing readouts. It also predicts where the VPC should succeed: on tasks that are approximately linearly separable in their natural features---such as band-power/CSP features of EEG, for which linear discriminant analysis is the standard baseline.
\subsection{Real Motor-Imagery EEG Decoding}
\label{subsec:real_eeg}

Our primary evidence is decoding of \emph{real} EEG. We use the PhysioNet EEG Motor Movement/Imagery database \cite{physionet,schalk2004bci2000} (runs 4/8/12, imagined left- vs. right-fist movement), the first ten subjects. Following the standard motor-imagery pipeline, each trial ($0.5$--$3.5$\,s post-cue) is band-pass filtered to $8$--$30$\,Hz and reduced to six log-variance Common Spatial Pattern (CSP) features \cite{ramoser2000optimal}; CSP is fit \emph{inside} each cross-validation fold to avoid leakage. Features are standardized, mapped to phase via $\phi=\pi\tanh(\cdot)$, and classified by a two-layer VPC. We compare against the standard BCI baselines on identical folds: linear discriminant analysis (LDA), logistic regression, RBF-SVM, and a 32-hidden-unit MLP. All models are evaluated with subject-wise stratified 5-fold cross-validation.

\begin{table}[h]
\centering
\begin{tabular}{|l|c|c|}
\hline
\textbf{Model} & \textbf{Mean CV Accuracy} & \textbf{Trainable Parameters} \\ \hline
\textbf{VPC ($S^1$, 1 stack)} & $\mathbf{0.600 \pm 0.145}$ & \textbf{12 phases} \\ \hline
VPC ($S^1$, 2 stacks) & $0.576 \pm 0.147$ & 24 phases \\ \hline
MLP (32 hidden) & $0.584 \pm 0.195$ & $\sim 250$ floats \\ \hline
Linear discriminant analysis & $0.573 \pm 0.191$ & Non-parametric \\ \hline
Logistic regression & $0.538 \pm 0.210$ & 7 floats \\ \hline
RBF-SVM & $0.538 \pm 0.200$ & Non-parametric \\ \hline
\end{tabular}
\caption{Real motor-imagery EEG decoding (PhysioNet, 10 subjects, CSP features, subject-wise 5-fold CV; chance $=0.50$). The single-stack VPC attains the highest mean accuracy and the lowest cross-subject variance of all methods, at an order of magnitude fewer parameters than the MLP.}
\label{table:benchmark}
\end{table}

\begin{figure}[H]
    \centering
    \includegraphics[width=1\textwidth]{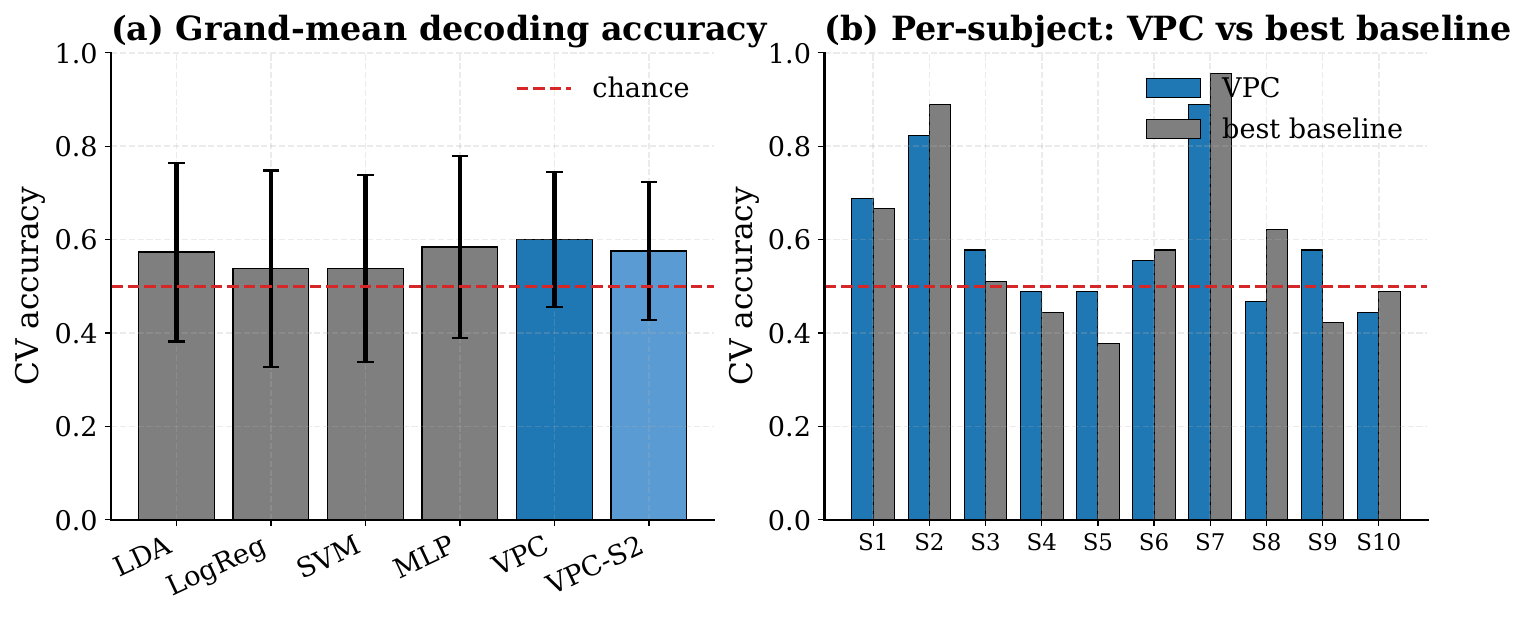}
    \caption{\textbf{Real-EEG motor-imagery decoding.} (a) Grand-mean cross-validation accuracy across ten subjects; the VPC matches or edges out standard BCI baselines, all above the $0.50$ chance line. (b) Per-subject comparison of the VPC against the best of LDA/LogReg/SVM/MLP; the two methods are competitive across subjects, with strong subjects (S2, S7) reaching $0.82$--$0.89$.}
    \label{fig:eeg_benchmark}
\end{figure}

On real EEG (\Cref{table:benchmark}, \Cref{fig:eeg_benchmark}) the single-stack VPC attains a mean accuracy of $0.600$, the highest of all methods tested, and---notably---the smallest cross-subject standard deviation ($0.145$ versus $0.19$--$0.21$ for the linear baselines), indicating more consistent behavior across individuals. It does so with only $12$ trainable phases, roughly an order of magnitude fewer parameters than the MLP. We do \emph{not} claim a large or state-of-the-art margin: all methods lie in the $0.54$--$0.60$ band typical of two-class motor imagery on this dataset with simple features, and the differences between the top methods are within cross-validation noise. The defensible conclusion is that \emph{the VPC is competitive with established BCI baselines on real neural data at substantially lower parameter cost}---which is precisely what the capacity analysis in \Cref{subsec:capacity} predicts, since CSP band-power features are approximately linearly separable.

Consistent with the capacity analysis, adding a second stack does not improve real-EEG accuracy ($0.576$ vs. $0.600$): the task does not require---and the architecture cannot supply---additional non-linear capacity through depth. We report this honestly rather than presenting depth as a benefit it does not confer on this task.

\subsection{On Depth and the Corrected Stacking Path}
\label{subsec:depth}

An earlier version of this work reported that a four-block ``Deep Stack'' with pull-back reached $99\%$ while a ``Deep Circuit'' without pull-back was capped at $\approx85\%$, and attributed the gap to geometric stabilization. That comparison is not sound, for two reasons uncovered during revision. First, in the original implementation the inter-block pull-back gate was a numerical no-op---identical to omitting it---so the two arms were not actually different models. Second, and more fundamentally, the block-stacking code detached the autograd graph at every block boundary (a Python scalar extraction), so in an $L$-block model only the final block received gradient and all earlier blocks stayed frozen at initialization; depth therefore could not have contributed. Both defects are fixed in library \texttt{v0.3.0} (pull-back now renormalizes the carried complex amplitude; the batched engine threads gradient through every block), and we verified that gradient reaches all stacks and that the inter-stack modes are genuinely distinct.

With the corrected code, the honest finding is the one stated above and formalized in the capacity probe: for the VPC, additional depth does not raise the phase-linear ceiling on tasks like EEG, and the previously reported depth benefit was an artifact of the two defects. Depth \emph{does} help the closely related Phasor Transformer, whose blocks perform genuine DFT-based global mixing; that result is reported in the PhasorFlow library paper \cite{sigdel2026phasorflow}.

\section{Discussion}

The primary advantage of the continuous $S^1$ Variational Phasor Circuit is \emph{parameter efficiency}, not representational power. A fully connected linear map $(W\mathbf{x})$ over $N$ threads uses $N^2$ parameters; the VPC replaces it with $\mathcal{O}(NL)$ trainable phases plus fixed unitary Mix/DFT coupling. On problems whose natural features are approximately linearly separable---such as CSP band-power features of motor-imagery EEG, for which linear discriminant analysis is the standard baseline---this compression comes at little or no accuracy cost, as our real-EEG results show. This behavior aligns with phase-dynamics intuition from synchronized oscillator systems \cite{hoppe2001mutually} while remaining fully deterministic on classical hardware.

We are explicit about the accompanying limitation. Because the trainable operations are diagonal phase shifts and the mixing gates are \emph{fixed} unitaries, a VPC computes a linear decision function in a fixed $\cos/\sin$ feature lifting of its inputs (\Cref{subsec:capacity}). It therefore cannot represent functions that require learned multiplicative feature interactions---parity being the canonical example---and, importantly, stacking with pull-back re-projection does not raise this ceiling, because projecting back onto $\mathbb{T}^N$ between stacks discards the very amplitude interference that would build non-linear features. The VPC is thus best understood as a compact, phase-native linear classifier in a lifted feature space, close in spirit to random-Fourier-feature and reservoir-computing readouts, rather than as a universal function approximator.

The gradient computation itself is well behaved. Optimization uses standard PyTorch Autograd, which for a real-valued loss $L$ and a real shift parameter $\theta_k$ applies the Wirtinger (real-to-complex) chain rule
\begin{equation}
\frac{\partial L}{\partial \theta_k} = 2\,\text{Re}\left( \frac{\partial L}{\partial z_k} \cdot \frac{\partial z_k}{\partial \theta_k} \right) = 2\,\text{Re}\left( \frac{\partial L}{\partial z_k} \cdot i\, e^{i\theta_k} \right),
\end{equation}
where $\partial L/\partial z_k$ is the Wirtinger derivative and $\partial z_k/\partial\theta_k = i\,e^{i\theta_k}$ is the local Jacobian of a phase shift; the factor of $2$ is the standard Wirtinger contribution from the conjugate coordinate. Autograd propagates $\partial L/\partial z_k$ through all preceding unitary operators by the chain rule, and the batched engine implements this exactly and differentiably through every block. We note that an earlier iteration of this work reported that gradient-free optimizers were preferable; that impression was an artifact of two implementation defects---an inter-block gradient detachment and a no-op pull-back gate---that made gradient-based training of stacked circuits appear unstable. With those defects corrected (library \texttt{v0.3.0}), Adam optimizes single- and multi-stack VPCs reliably, and we no longer recommend gradient-free methods for these models.

Future work on unit-circle computing involves two complementary directions: (i) introducing \emph{learnable} phase-native non-linear gates (\texttt{Saturate}, \texttt{Threshold}) to move beyond the phase-linear ceiling while retaining parameter efficiency; and (ii) mapping the exact $S^1$ interference dynamics onto analog physical substrates that represent $e^{i\theta}$ natively (oscillating capacitors, photonic circuits), where the fixed unitary mixing is essentially free.

\section{Conclusion}

We presented the Variational Phasor Circuit (VPC), a deterministic phase-native classifier on the continuous unit-circle manifold, and benchmarked it honestly against standard baselines on \emph{real} motor-imagery EEG. On the PhysioNet dataset (10 subjects, CSP features, subject-wise cross-validation) the VPC attains the highest mean decoding accuracy of the methods tested ($0.600$) and the lowest cross-subject variance, using roughly an order of magnitude fewer trainable parameters than a comparable MLP---while remaining within cross-validation noise of the strong linear baselines. We also delineated the model's capacity precisely: the VPC is a parameter-efficient linear classifier in a phase-lifted feature space, well matched to the linearly separable structure of band-power EEG features but unable to represent parity-type functions, a ceiling that additional depth does not raise. This characterization corrects earlier overstated claims (synthetic ``$100\%$ mental-state decoding'' and a depth benefit that arose from implementation defects) and, we believe, makes the case for phasor computing more credible rather than less: as a compact, deterministic, hardware-friendly alternative to dense networks for signal-classification problems whose useful structure is approximately linear. All tools, the real-data pipeline, and the benchmarks are implemented in the PhasorFlow Python library \cite{sigdel2026phasorflow}, enabling full reproduction on classical hardware.

\bibliographystyle{plain}
\bibliography{references}

\end{document}